\documentclass{article}

\PassOptionsToPackage{numbers, sort&compress}{natbib}

\usepackage[main, preprint]{neurips_2026}

\usepackage[utf8]{inputenc} 
\usepackage[T1]{fontenc}    
\usepackage[hidelinks]{hyperref}       
\usepackage{url}            
\usepackage{booktabs}       
\usepackage{nicefrac}       
\usepackage{microtype}      
\usepackage[dvipsnames]{xcolor}         
\usepackage{rotating}

\usepackage{array}
\usepackage{adjustbox}
\usepackage{graphicx}
\usepackage{multirow}
\usepackage{bm}
\usepackage{threeparttable}

\renewcommand{\arraystretch}{1.1}

\usepackage{amsmath}
\usepackage{amssymb}
\usepackage{tabularx}

\newcolumntype{C}{>{\centering\arraybackslash}X}
\newcolumntype{R}{>{\raggedleft\arraybackslash}X}

\newcommand{\ra}[1]{\renewcommand{\arraystretch}{#1}}

\newcommand{\best}[1]{%
\textbf{\boldmath #1}
}

\newcommand{\res}[2]{\ensuremath{#1 \pm #2}}
\newcommand{\rankres}[2]{\ensuremath{\bm{#1 \pm #2}}}

\title{Scalable Memristive-Friendly Reservoir Computing for Time Series Classification}

\author{
Coşku Can Horuz \\
Adaptive AI Research Group \\
University of Lübeck, Germany\\
\texttt{cosku.horuz@uni-luebeck.de}
\And
Andrea Ceni \\
Department of Computer Science \\
 University of Pisa, Italy\\
\texttt{andrea.ceni@unipi.it}
\And
Claudio Gallicchio \\
Department of Computer Science \\
University of Pisa, Italy\\
\texttt{claudio.gallicchio@unipi.it}
\And
Sebastian Otte \\
Adaptive AI Research Group \\
University of Lübeck, Germany\\
\texttt{sebastian.otte@uni-luebeck.de}
}

\begin{document}

\maketitle

\begin{abstract} 
Memristive devices present a promising foundation for next-generation information processing by combining memory and computation within a single physical substrate. This unique characteristic enables efficient, fast, and adaptive computing, particularly well suited for deep learning applications. Among recent developments, the memristive-friendly echo state network (MF-ESN) has emerged as a promising approach that combines memristive-inspired dynamics with the training simplicity of reservoir computing, where only the readout layer is learned. Building on this framework, we propose memristive-friendly parallelized reservoirs (MARS), a simplified yet more effective architecture that enables efficient scalable parallel computation and deeper model composition through novel subtractive skip connections. This design yields two key advantages: substantial training speedups of up to 21x over the inherently lightweight echo state network baseline and significantly improved predictive performance. Moreover, MARS demonstrates what is possible with parallel memristive-friendly reservoir computing: on several long sequence benchmarks our compact gradient-free models substantially outperform strong gradient-based sequence models such as LRU, S5, and Mamba, while reducing full training time from minutes or hours down seconds or even only a few hundred milliseconds. Our work positions parallel memristive-friendly computing as a promising route towards scalable neuromorphic learning systems that combine high predictive capability with radically improved computational efficiency, while providing a clear pathway to energy-efficient, low-latency implementations on emerging memristive and in-memory hardware.
\end{abstract}

\section{Introduction}
The advancement of deep learning has led to transformative progress across numerous scientific and industrial fields. Yet, the increasing scale of modern workloads has amplified concerns regarding computational efficiency, scalability, and energy consumption \citep{bolon2024review}. In particular, growing reliance on transformer-based large architectures has increased the need for alternative models that are both computationally efficient and designed to meet emerging hardware constraints. In response, paradigms such as neuromorphic computing \cite{schuman2022opportunities} and reservoir computing (RC) \cite{nakajima2021reservoir,tanaka2019recent} have garnered significant interest. These frameworks aim to emulate the efficiency of biological or physical systems while enabling effective temporal signal processing.
Within this domain, echo state networks (ESNs) \cite{jaeger2001,lukovsevivcius2012practical}, a reservoir computing variant of recurrent neural networks (RNNs), have long served as a lightweight and energy-efficient class of recurrent models for time-series classification. Building on this foundation, the memristive-friendly echo state network (MF-ESN) \cite{pistolesi2025memristive} has been introduced to model the dynamics of memristive systems, which are nanoscale devices exhibiting nonlinear fading memory \cite{miranda2020modeling,milano2022materia}. MF-ESN offers a software abstraction of memristive dynamics while preserving the training simplicity of classical ESNs, making it a suitable candidate for low-power machine learning applications.
Despite these advantages, ESNs and MF-ESNs are inherently sequential and poorly suited for parallel execution on modern hardware accelerators. This limitation has contributed to the dominance of transformers, whose attention mechanism is highly parallelizable. In contrast, recent advances in structured recurrent models, such as S4 \cite{gu2022efficiently}, LRU \cite{orvieto2023resurrecting}, Mamba \cite{gu2024mamba}, and others \cite{de2024griffin,beck2024xlstm}, demonstrate that RNNs can also be formulated for efficient parallelism.

In this work, we present a parallel implementation of MF-ESN model by leveraging the parallel scan algorithm \cite{blelloch1990prefix}. Our approach enables scalable forward computation and efficient stacking of memristive layers. Experiments show substantial speedups and improved performance over baseline reservoir models as well as against large scale gradient-based state-of-the-art (SoTA) architectures demonstrating the viability of combining neuromorphic efficiency with modern parallel computing.

\section{Background}\label{sec:background}
Understanding the context and foundations of this work requires considering three key domains: reservoir computing, memristive-inspired dynamics, and recent advances in parallel recurrent computation.

\paragraph{\textbf{Reservoir Computing.}}
RC is a computational framework for processing sequential data, introduced as a lightweight alternative to fully trained RNNs. Traditional RNNs face significant challenges during training, especially when learning long-term dependencies, due to the vanishing and exploding gradient problem \cite{pascanu2013difficulty}. RC addresses these issues by using a large, fixed reservoir of randomly connected nonlinear units. The input signal drives the reservoir, which produces a dynamic, high-dimensional echo of past inputs. Crucially, the internal weights of the reservoir remain untrained; only a simple readout layer is optimized, making the learning process significantly efficient and stable. Echo state networks, introduced by Jaeger in 2001 \cite{jaeger2001}, are among the most widely studied models in reservoir computing. Formally, given input $\mathbf{x}_t \in \mathbb{R}^{N_x}$, the \textit{leaky} ESN state $\mathbf{h}_t \in \mathbb{R}^{N_h}$ is updated as:
\begin{equation}
\mathbf{h}_{t+1} = (1 - \alpha) \mathbf{h}_t + \alpha \tanh(\mathbf{W}^{x} \mathbf{x}_{t+1} + \mathbf{W}^{h} \mathbf{h}_t + \mathbf{b})
\end{equation}
where $\alpha$ is the leaking rate, $\mathbf{W}^{x}$ is the input matrix, $\mathbf{W}^{h}$ is the recurrent matrix, and $\mathbf{b}$ is the bias vector. The output is computed as:
\begin{equation}
\mathbf{y}_{t+1} = \mathbf{W}^{y} \mathbf{h}_{t+1}
\end{equation}
The output weight tensor $\mathbf{W}^{y}$ is the only trainable component of the network, typically optimized with global linear (ridge) regression/classification. The reservoir is usually scaled to operate near the so-called edge of stability, ensuring that both fading memory and rich dynamics are captured in its activations.
This framework presents a more efficient path compared to standard backpropagation through time, bypassing expensive and often unstable gradient computations.

In particular, ESNs implement a sparsely connected random RNN whose internal weights remain fixed post-initialization, with training focused solely on the linear readout. Over time, ESNs have proven effective for a range of tasks and attracted interest for hardware-efficient implementations in neuromorphic and low-power computing platforms \cite{zhang2023survey}.

\paragraph{\textbf{Memristive Dynamics.}}
Memristive computation refers to the use of memristors, which are electronic components whose resistance depends on the history of voltage or current. Due to their inherent nonlinearity and memory retention, memristors have been explored as building blocks for neuromorphic computing systems.
Two recent studies propose recurrent neural architectures that incorporate neuron models inspired by the physical dynamics of memristive nanowires \cite{pistolesi2025esannmemristive,pistolesi2025memristive}. These models rely on a nonlinear update rule derived from a potentiation–depression rate balance, describing how a neuron's internal state (analogous to conductance) evolves in response to input stimuli. The potentiation and depression rates are defined as follows:
\begin{align}
    K_{p}(\mathbf{x}) &= K_{p_0} \cdot e^{\eta_{p}\mathbf{x}} \\
    K_{d}(\mathbf{x}) &= K_{d_0} \cdot e^{-\eta_{d}\mathbf{x}}
\end{align}
$K_{p_0}$, $\eta_{p}$, $K_{d_0}$ and $\eta_{d}$ 
are the memristive-based parameters that are determined by their physical plausibility, namely, $K_{p_{0}} = 0.0001$, $\eta_{p} = 10$, $K_{d_{0}} = 0.5$, and $\eta_{d} = 1$ \cite{milano2022connectome}. These two equations model exponential growth/decay behavior and, in a way, compete with each other. Especially the term $e^{10}$ in potentiation, might dominate the memristive dynamics rapidly.

The time-discretized sequential equations of a memristive-friendly network are defined as follows:
\begin{align}
    \mathbf{z}_{t+1} &= \operatorname{RESCALE}(\mathbf{W}^{h}\mathbf{h}_{t} + \mathbf{W}^{x}\mathbf{x}_{t+1} + \mathbf{b})\label{eq:z}\\
    \mathbf{q}_{t+1} &= (K_{p}(\mathbf{z}_{t+1}) + K_{d}(\mathbf{z}_{t+1})) \odot \mathbf{h}_{t}\\
    \mathbf{r}_{t+1} &= K_{p}(\mathbf{z}_{t+1}) - \mathbf{q}_{t+1}\\
    \mathbf{h}_{t+1} &= \mathbf{r}_{t+1}\Delta + \gamma \mathbf{h}_{t}
\end{align}
where $\mathbf{W}^{h} \in \mathbb{R}^{N_{h} \times N_{h}}$ is the reservoir recurrent weight matrix, $\mathbf{W}^{x} \in \mathbb{R}^{N_{h} \times N_{x}}$ is the input matrix, $\mathbf{b} \in \mathbb{R}^{N_{h}}$ is the bias vector, and $\odot$ denotes the Hadamard product. The $\operatorname{RESCALE}$ function is applied pointwise and it constrains the activations such that they stay in a stable range. It also prevents the exploding dynamics by the exponentiations in $K_{d}$ and $K_{p}$. It is defined as
\begin{equation}
    \operatorname{RESCALE}(z) = \frac{b - a}{1 + \operatorname{exp}(-zs)} + a
\end{equation}
where $a = 0.35$ and $b = 1.15$ are fixed parameters as in \cite{pistolesi2025memristive}, whereas $s$ is a hyperparameter that is adapted to the problem. $\operatorname{RESCALE}$ is a modified version of a sigmoid function which is bounded by the values $a$ and $b$. $s$ decides for the steepness of the nonlinearity in between the boundaries. A higher value of $s$ induces a stronger degree of non-linearity.

\paragraph{\textbf{Parallelized Linear Recurrent Computation.}}
The ongoing revival of RNNs has started with the realization that linear recurrences embedded within a deep non-linear architecture are actually quite expressive \cite{gu2022efficiently,orvieto2023resurrecting,gu2024mamba,feng2024were}. A linear recurrent mapping of the form $\mathbf{h}_{t+1} = \mathbf{W}^{h}\mathbf{h}_{t} + \mathbf{W}^{x}\mathbf{x}_{t+1}$ can be realized with global convolution via fast Fourier transform, computing the full hidden state sequence at once \cite{gu2021combining}. It is also possible to perform this process using parallel scan, occasionally called the prefix sum \cite{martin2018parallelizing,qin2023hierarchically}. The computational gain is particularly pronounced, when $\mathbf{W}^h$ is structured \cite{gu2022efficiently}. Especially in the optimally structured case, namely, when $\mathbf{W}^h$ is a diagonal matrix, the recurrent computation is tremendously efficient on a parallel processor and still expressive enough \cite{gu2022diagonal}.

Generally, it is possible to apply the parallel scan algorithm for the following equation:
\begin{equation}\label{eq:lin_rec}
    \mathbf{h}_{t+1} = \mathbf{a}_{t+1} \odot \mathbf{h}_{t} + \mathbf{b}_{t+1}
\end{equation}
as long as $\odot$ and $+$ are binary associative operators. In our work, we use a specific application of parallel scan that maps the equation into the log-space to be able to calculate the product with the cumulative sum \cite{heinsen2023parallelization}:

\begin{equation}
\log \mathbf{h}_{t+1} = \log \mathbf{a}_{t+1} + \log \mathbf{h}_t + \log\left(1 + \exp(\log \mathbf{b}_{t+1} - \log \mathbf{a}_{t+1} - \log \mathbf{h}_t)\right)
\end{equation}

By computing cumulative sum in the log-domain, the entire sequence $\{\mathbf{h}_t\}_{t=1}^T$ can be computed in parallel. Let:
\begin{equation}
\boldsymbol{\alpha}_t := \sum_{i=1}^{t} \log \mathbf{a}_i
\quad \text{and} \quad
\boldsymbol{\beta}_t := \log \mathbf{b}_t.
\end{equation}
Then the parallel scan computes
\begin{equation}
\log \mathbf{h}_t = 
\boldsymbol{\alpha}_t + \log \left( \mathbf{h}_{0}
+ \sum_{i=1}^{t} \exp(\boldsymbol{\beta}_i - \boldsymbol{\alpha}_i) \right)
\end{equation}
and we recover the final output as
\begin{equation}
\mathbf{h}_t = \exp\left( \boldsymbol{\alpha}_t + \log
\left(\mathbf{h}_{0} + 
\sum_{i=1}^{t} \exp(\boldsymbol{\beta}_i - \boldsymbol{\alpha}_i) \right) \right)
\end{equation}
 
This log-space computation, however, comes with a caveat: the inputs of the logarithm (i.e. $\mathbf{a}$ and $\mathbf{b}$) must be positive if the computation must stay in the real domain. Nonetheless, this is not a problem in our case as $K_{d}$ and $K_{p}$ are always positive due to the exponentiation. 

\section{Memristive-Friendly Parallelized Reservoir (MARS)}\label{sec:mars}
We propose a parallelization scheme for the memristive dynamics described in \autoref{sec:background}. We denote the proposed model as memristive-friendly parallelized reservoir (MARS).

Only linear recurrent relations can be parallelized. Since $\operatorname{RESCALE}$ is non-linear, it is not possible to compute $\mathbf{z}$ from \autoref{eq:z} in parallel. Therefore, we set $\mathbf{W}^{h} = \mathbf{0}$.
Accordingly, the equation reduces to
\begin{equation}
    \mathbf{z}_{[1:T]} = \operatorname{RESCALE}(\mathbf{W}^{x}\mathbf{x}_{[1:T]} + \mathbf{b})
\end{equation}

In this form, it is possible to apply the weighted sum and $\operatorname{RESCALE}$ for the whole sequence (denoted by $[1:T]$) at once. After the redefinition of $\mathbf{z}$, the sequential equations can be reformulated as follows:
\begin{align}
    \mathbf{h}_{t+1} &= \mathbf{r}_{t+1}\Delta + \gamma\mathbf{h}_{t} \label{eq:par_mem}\\
    &= \left( K_{p}(\mathbf{z}_{t+1}) - \mathbf{q}_{t+1} \right)\Delta + \gamma\mathbf{h}_{t}\nonumber\\
    &= \left( K_{p}(\mathbf{z}_{t+1}) - \left[ K_{p}(\mathbf{z}_{t+1}) + K_{d}(\mathbf{z}_{t+1}) \right] \odot\mathbf{h}_{t} \right)\Delta + \gamma\mathbf{h}_{t}\nonumber\\
    &= \Delta K_{p}(\mathbf{z}_{t+1}) - \Delta \left[ K_{p}(\mathbf{z}_{t+1}) + K_{d}(\mathbf{z}_{t+1}) \right] \odot\mathbf{h}_{t} + \gamma\mathbf{h}_{t}\nonumber\\
    &= \left( \gamma - \Delta \left[ K_{p}(\mathbf{z}_{t+1}) + K_{d}(\mathbf{z}_{t+1}) \right] \right) \odot\mathbf{h}_{t} + \Delta K_{p}(\mathbf{z}_{t+1})\nonumber
\end{align}

The final formulation has exactly the same form as \autoref{eq:lin_rec}, which allows the employment of parallel scan algorithm. Consequently, $\mathbf{a}_{t+1} := \left( \gamma - \Delta \left[ K_{p}(\mathbf{z}_{t+1}) + K_{d}(\mathbf{z}_{t+1}) \right] \right)$ and $\mathbf{b}_{t+1} := \Delta K_{p}(\mathbf{z}_{t+1})$. Note that it is possible to compute $\mathbf{a}_{[1:T]}$ and $\mathbf{b}_{[1:T]}$ at once as they encode no recurrent relations. Computing these two terms and mapping them into the log-space yields a formulation that is well suited for parallel scan, with $\mathbf{a}_{[1:T]}$ and $\mathbf{b}_{[1:T]}$ as inputs. MARS preserves the memristive dynamics of MF-ESN, differing only in the absence of recurrent weights.

\begin{figure}[b]
\centering
\includegraphics[width=0.7\textwidth]{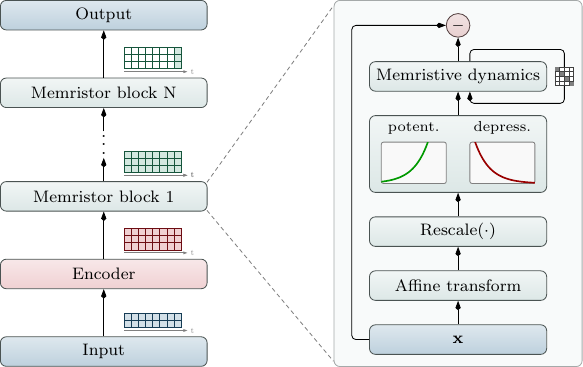}
\caption{Abstract depiction of the MARS architecture. Stacked memristor blocks include linear operations, parallel recurrent computation and subtractive skip connections. Solely the last time step hidden state of the last layer is used for output mapping by ridge classification.} \label{fig:mars}
\end{figure}

\begin{figure}[ht]
\includegraphics[width=\textwidth]{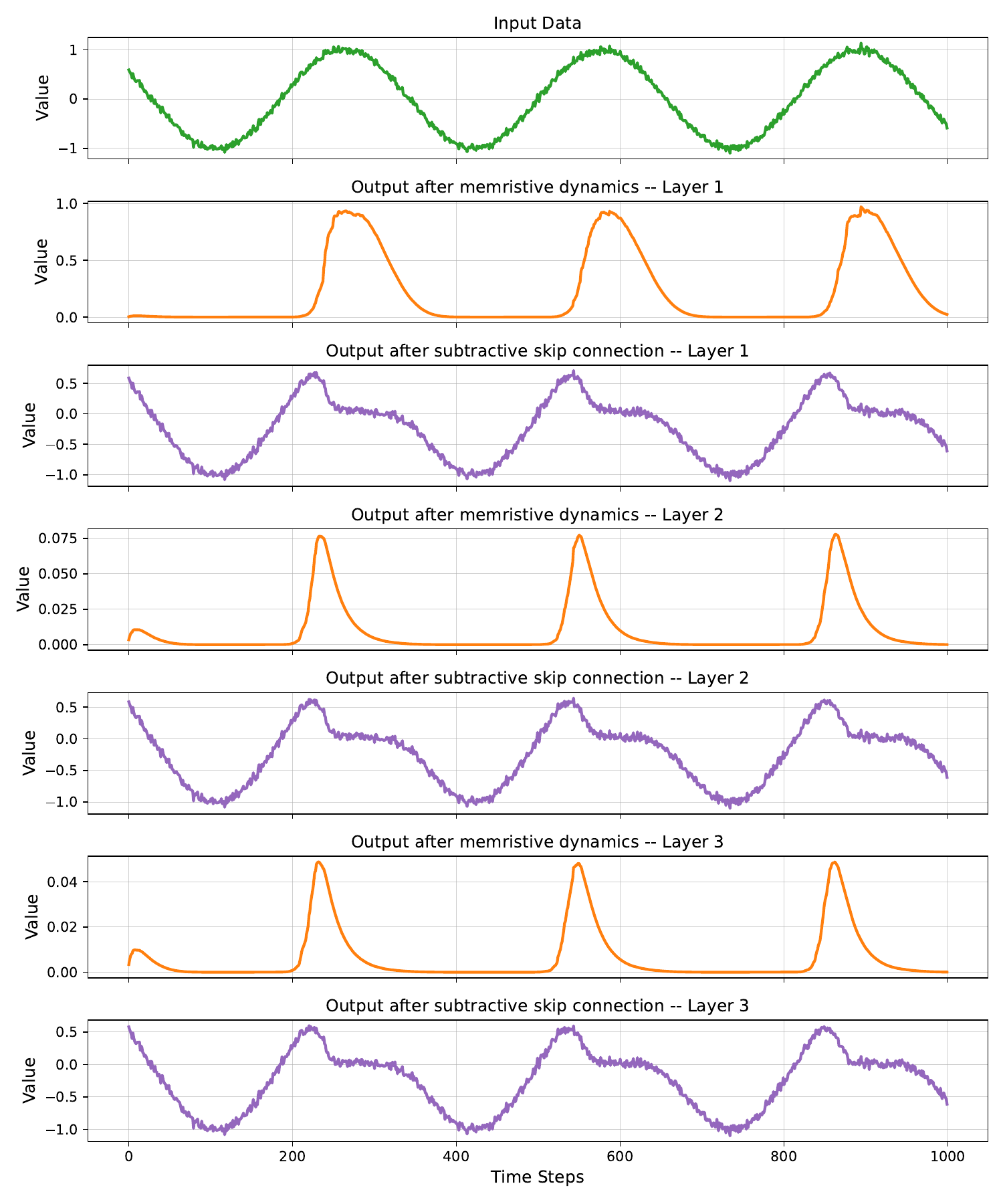}
\caption{A toy example illustrating the low-pass filtering effect of subtractive skip connections combined with memristive dynamics. The signal contains higher frequencies in deeper layers.} \label{fig:hiddens}
\end{figure}

Parallel computation allows a model to be efficient and scalable but it is not the only contribution of the recent linear recurrent models. Integrating stacked recurrent layers into a coherent architecture is also as important. MARS benefits significantly from depth, as stacking enhances expressivity despite the linear temporal propagation. \autoref{fig:mars} shows our proposed MARS architecture. First the input is mapped to the hidden state via an encoder. Afterwards, memristive blocks are applied for $N$ layers. Linear operations such as affine transformation, $\operatorname{RESCALE}$, potentiation and depression are applied to complete sequences. The recurrent memristive computation is performed using a diagonal recurrent matrix as shown in \autoref{eq:par_mem}. The previous layer information flows through a skip connection. Usually skip connection is applied by adding the input from the previous layer in order to foster a smooth gradient flow \cite{he2016deep}. However, as MARS is a reservoir network, there is no gradient computation. Skip connections serve a different purpose in this case. Memristive computation is substantially different than vanilla recurrent computation as it is constraint by the physical device dynamics. Potentiation and depression are exponentially growing and decaying terms, respectively. Especially the potentiation can become intractable quickly. Therefore, the memristive dynamics fall into a certain regime (decay or growth) and they encode low-frequency information. Subtracting this information creates a filtering effect. We call this operation \textit{subtractive skip connection}. \autoref{fig:hiddens} is a visual demonstration of this filtering effect. In a completely clean setting with no affine transformation and rescaling, a single sinusoidal signal with a marginal noise level is fed into the memristive layer. The output is merged into the input with a subtractive skip connection. After three layers, the resulting signal contains more high-frequency components.

\section{Experiments}
We have conducted experiments in two different scenarios. First, we measure the scaling behavior and classification ability of MARS as opposed to the classical leaky ESN and MF-ESN. The first experimental setup is designed to assess the improvements of our model over previous versions. Second, we test MARS in several real-world classification benchmarks against heavyweight gradient-based SoTA models in order to show that efficiency does not mean poor performance.

\subsection{MARS within reservoir computing}\label{sec:mars_in_res}
\begin{figure}[ht]
\includegraphics[width=\textwidth]{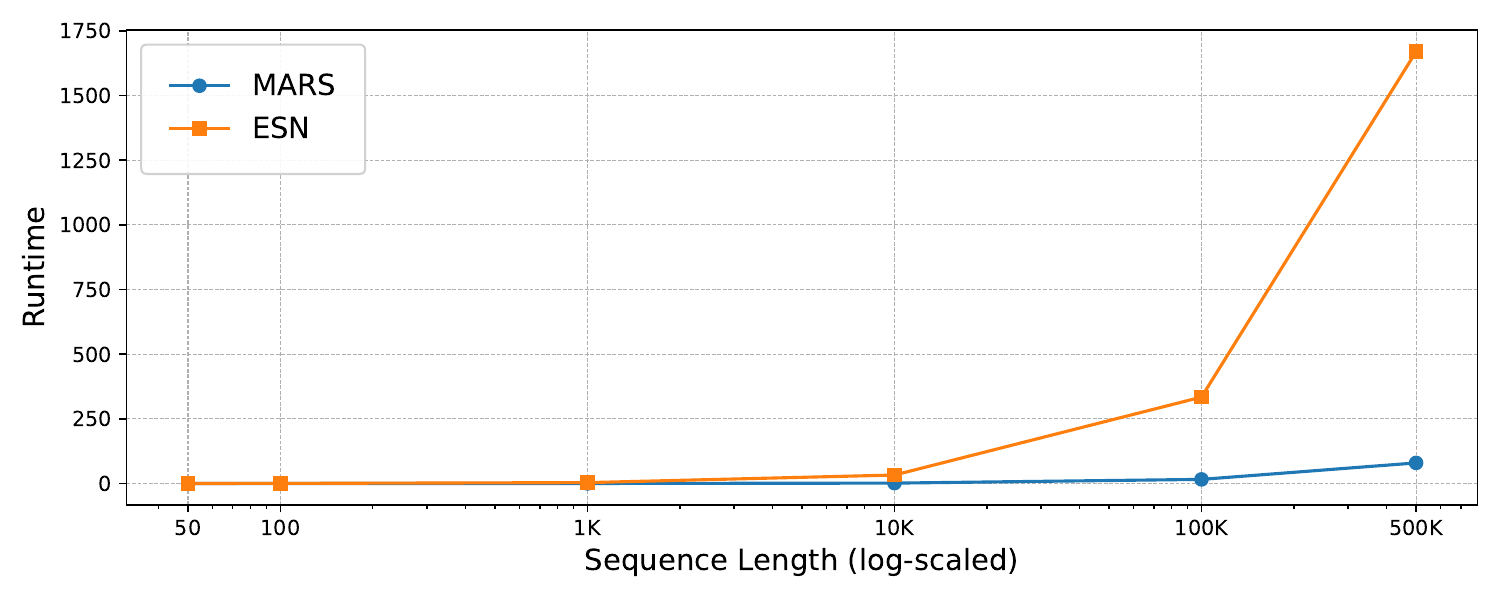}
\caption{Runtime comparison of MARS and ESN across varying sequence lengths. The x-axis (log-scaled) represents the input sequence in time steps, ranging from $50$ to $500k$. The y-axis shows the total runtime in seconds required to forward process each sequence length over $100$ iterations with the batch size of $10$. MARS demonstrates significantly better scalability with near-logarithmic complexity, while ESN exhibits a steep increase in processing time as the input size grows.} \label{fig:runtime}
\end{figure}

\paragraph{Runtime Analysis.}
The exclusion of the recurrent weight matrix $\mathbf{W}^h$ and the parallel computation along the temporal axis makes MARS much more resource-efficient. In this set of experiments, we investigate the runtime difference with the classical leaky ESN evaluated by wall-clock time. We have created randomly initialized input tensors in range $[0, 1]$ which vary in sequence length. We did not include the optimization of the output layer as it should take the same time for both models. Also, the number of trainable and fixed parameters are roughly equal because MARS has $3$ layers as opposed to the single layer in ESN. The results shown in \autoref{fig:runtime} and \autoref{tab:runtime} represent the same experiments and clearly show the logarithmic parallel complexity of MARS as opposed to the linear complexity of ESNs. We emphasize that, here, MARS is compared against a model that is already considered a highly efficient, lightweight network.
\begin{table}[!ht]
\caption{Forward pass runtime analysis of ESN and MARS with different length inputs. The results are given in seconds and obtained from the same experiments as those shown in \autoref{fig:runtime}. Nvidia GeForce RTX 4090 GPU is used for the experiments. For an input length of $500k$, MARS requires $1.3$ minutes, compared to $27.8$ minutes for ESN.}
\footnotesize
\label{tab:runtime}
\ra{1.2}
\begin{tabularx}{\linewidth}{XCCCCCC}
\toprule
\textbf{Model} & $50$ & $100$ & $1k$ & $10k$ & $100k$ & $500k$ \\
\midrule
\textbf{MARS} & $0.0995$ & $0.1012$ & $0.1622$ & $1.5079$ & $15.9531$ & $79.6439$ \\
\textbf{ESN} & 0.2169 & $0.3835$ & $3.4795$ & $32.7758$ & $334.0719$ & $1,669.3938$ \\
\bottomrule
\end{tabularx}
\end{table}

\paragraph{Time Series Classification.} We benchmark MARS against classical ESN and MF-ESN on multiple short-length time series classification tasks from the University of California, Riverside (UCR) archive \cite{dau2019ucr}. UCR includes datasets of diverse nature, some originating from real-world scenarios. Statistics of the selected datasets can be found in \autoref{tab:datasets}. Coffee dataset has the longest sequence with $286$ time steps. 

All the compared models have roughly the same number of learnable and randomly initialized fixed parameters. The results for ESN and MF-ESN are taken directly from \cite{pistolesi2025memristive} and are reported in \autoref{tab:accuracy}. The hyperparameters of MARS, which include input scaling, bias scaling, $\Delta$, $\gamma$ and $s$, were selected using a procedure similar to that described in \cite{pistolesi2025memristive}. In \autoref{tab:hypers} we provide the best hyperparameters found for our proposed model for each dataset.

Albeit its exceptional efficiency and the linear recurrent nature, MARS outperforms its competitors in all tasks except one. This outstanding performance can be attributed to the combination of depth and nonlinear transformations introduced by $\operatorname{RESCALE}$ in between layers. While each individual recurrence is linear (as $\mathbf{W}^{h} = 0$) and therefore limited in expressiveness, stacking multiple such layers interleaved with nonlinearities significantly enhances the model’s capacity to capture complex dynamics. This design reflects recent progress in sequential modeling, such as S4 \cite{gu2022efficiently}, LRU \cite{orvieto2023resurrecting}, and Mamba \cite{gu2024mamba}, which demonstrate that deep compositions of linear state-space recurrent dynamics and nonlinearities can effectively capture complex temporal patterns. These results suggest that, efficient linear temporal computation in a network can compensate for, or even outperform, traditional recurrent connections within a reservoir.
\begin{table}[t!]
\centering
\caption{Test accuracy statistics for seven different time series classification tasks. The results of ESN and sequential MF-ESN~is taken from~\cite{pistolesi2025memristive}. MARS column shows our results. \best{Best models} are  made salient.}
\label{tab:accuracy}
\footnotesize
\ra{1.2}
\begin{tabularx}{0.8\linewidth}{XCCC}
\toprule
\textbf{Task} & ESN & MF-ESN & MARS \\
\midrule
\textbf{Epilepsy} & $0.868\pm0.015$ & \best{$0.961\pm0.007$} & $0.956\pm0.008$ \\
\textbf{SyntheticControl} & $0.893\pm0.010$ & $0.950\pm0.007$ & \best{$0.980\pm0.004$} \\
\textbf{GunPoint} & $0.593\pm0.023$ & $0.769\pm0.070$ & \best{$0.956\pm0.003$} \\
\textbf{ECG5000} & $0.915\pm0.002$ & $0.924\pm0.002$ & \best{$0.939\pm0.001$} \\
\textbf{Coffee} & $0.586\pm0.029$ & $0.586\pm0.017$ & \best{$1.000\pm0.000$} \\
\textbf{JapaneseVowels} & $0.984\pm0.003$ &  $0.974\pm0.004$ & \best{$0.989\pm0.003$} \\
\textbf{Wafer} & $0.986\pm0.002$ & $0.986\pm0.001$ & \best{$0.996\pm0.001$} \\
\bottomrule
\end{tabularx}
\end{table}

\subsection{MARS within gradient-based SoTA models}
We evaluate MARS with large gradient-based models, that set state-of-the-art performance for some classification tasks. It is rather an unusual approach to compare a reservoir network in this setting. Gradient signals are typically highly adaptive and in large scale the optimization can reach well generalizing local minima. Therefore, no-gradient models are not expected to achieve comparable performance, particularly given that they are faster by several orders of magnitude. We consider a line of work that focuses on recent recurrent architectures such as state space models. \citet{walker2024logncde} have recently introduced a log-transformed neural controlled differential equation model and compared it with latest recurrent models such as S4 \cite{gu2022efficiently}, S5 \cite{smith2023simplified}, Mamba \cite{gu2024mamba} and LRU \cite{orvieto2023resurrecting}. Their benchmark consists of six real-world datasets from the University of East Anglia (UEA) Multivariate Time Series Classification Archive (UEA-MTSCA) \cite{bagnall2018uea}. The longest sequences were chosen specifically for increased difficulty. One year after, \citet{rusch2025oscillatory} took the exact experimental design and benchmarked their linear oscillatory state space approach against the reported results by \citet{walker2024logncde}. We continue this journey but with a substantially different approach. This time the proposed model uses no gradients and is tiny in comparison. We follow the exact experimental setting as depicted in \cite{walker2024logncde} including the data splitting strategy, removal of repeated sequences and seeds. The only difference is the Python framework. In contrast to the previous works, our code is written in PyTorch \cite{paszke2019pytorch} instead of Jax \cite{jax2018github}.

\begin{table}[t!]
\caption{Classification results depicted by test accuracies averaged over 5 runs on selected long-sequence real-world datasets. All models are optimized based on the same training protocol for a fair comparison. The dataset names are abbreviations of the following UEA-MTSCA datasets: EigenWorms (Worms), SelfRegulationSCP1 (SCP1), SelfRegulationSCP2 (SCP2), EthanolConcentration (Ethanol), Heartbeat, MotorImagery (Motor). All results apart from MARS are taken from \cite{rusch2025oscillatory}. \best{Best models} are  made salient.}
\footnotesize
\label{tab:long_benchmarks}
\ra{1.2}
\begin{tabularx}{\linewidth}{Xcccccc}%
\toprule
 & Worms & SCP1 & SCP2 & Ethanol & Heartbeat & Motor \\
Seq.\ length & 17{,}984 & 896 & 1{,}152 & 1{,}751 & 405 & 3{,}000 \\
\#Classes     & 5 & 2 & 2 & 4 & 2 & 2  \\
\midrule
NRDE
  & \res{83.9}{7.3}
  & \res{80.9}{2.5}
  & \res{53.7}{6.9}
  & \res{25.3}{1.8}
  & \res{72.9}{4.8}
  & \res{47.0}{5.7} \\
NCDE
  & \res{75.0}{3.9}
  & \res{79.8}{5.6}
  & \res{53.0}{2.8}
  & \res{29.9}{6.5}
  & \res{73.9}{2.6}
  & \res{49.5}{2.8} \\
Log-NCDE
  & \res{85.6}{5.1}
  & \res{83.1}{2.8}
  & \res{53.7}{4.1}
  & \res{34.4}{6.4}
  & \res{75.2}{4.6}
  & \res{53.7}{5.3} \\
LRU
  & \res{87.8}{2.8}
  & \res{82.6}{3.4}
  & \res{51.2}{3.6}
  & \res{21.5}{2.1}
  & \rankres{78.4}{6.7}
  & \res{48.4}{5.0} \\
S5
  & \res{81.1}{3.7}
  & \res{89.9}{4.6}
  & \res{50.5}{2.6}
  & \res{24.1}{4.3}
  & \res{77.7}{5.5}
  & \res{47.7}{5.5} \\
S6
  & \res{85.0}{16.1}
  & \res{82.8}{2.7}
  & \res{49.9}{9.4}
  & \res{26.4}{6.4}
  & \res{76.5}{8.3}
  & \res{51.3}{4.7} \\
Mamba
  & \res{70.9}{15.8}
  & \res{80.7}{1.4}
  & \res{48.2}{3.9}
  & \res{27.9}{4.5}
  & \res{76.2}{3.8}
  & \res{47.7}{4.5} \\
LinOSS-IMEX
  & \res{80.0}{2.7}
  & \res{87.5}{4.0}
  & \rankres{58.9}{8.1}
  & \res{29.9}{1.0}
  & \res{75.5}{4.3}
  & \res{57.9}{5.3} \\
LinOSS-IM
  & \rankres{95.0}{4.4}
  & \res{87.8}{2.6}
  & \res{58.2}{6.9}
  & \res{29.9}{0.6}
  & \res{75.8}{3.7}
  & \rankres{60.0}{7.5} \\
\midrule
\textbf{MARS}
  & \res{71.11}{5.41}
  & \rankres{91.76}{2.76}
  & \res{50.69}{3.78}
  & \res{34.68}{6.67}
  & \res{74.19}{5.93}
  & \res{55.79}{7.59} \\
\textbf{MARS-TC}
  & \res{72.22}{3.40}
  & \res{89.65}{4.51}
  & \res{53.10}{9.24}
  & \rankres{37.97}{6.52}
  & \res{72.90}{8.41}
  & \res{55.79}{7.17} \\
\bottomrule
\end{tabularx}
\end{table}

Considering the long sequences in this benchmark, fixed parameters of the encoder might not be able to capture the temporal dependencies. Therefore, we introduce another version of MARS in which we apply 1D temporal convolution (TC) \cite{kalchbrenner2016neural} layer on the input before the encoding layer. This operation combines local temporal information via convolution kernels that are randomly initialized and remain fixed during training. Irrespective of the input dimension, the convolution layer maps the input to $20$ channels in all tasks. A linear encoder layer is still applied afterwards. We denote this version as MARS-TC in the results.

The results given in \autoref{tab:long_benchmarks} show a remarkable trend. Instead of hours of training over many iterations, it is possible to reach challenging, or even better, performance with the memristive-inspired MARS. Although not always setting the SoTA, MARS obtains competitive results in almost all tasks. EigenWorms dataset requires a specific attention as it is approximately $18k$ long. As analyzed in \autoref{sec:mars}, the memristive hidden state dynamics fall into two different regimes (potentiation or depression) depending on the input. This regime decision happens quickly and for extremely long sequences, it leads to information loss. We attribute the dropped performance of MARS to this inherent phenomenon of a memristive memory system.

The simplicity of MARS is further evident in wall-clock runtime results and the number of trainable parameters (see \autoref{tab:training_time} and \ref{tab:params}). In comparison to gradient-based models, MARS trains only the output layer. Still, MARS is a deep network with layer numbers changing between 2 to 10. The parameters across layers are initialized only in the beginning and are not learned. Reservoir networks are trained in a single run with no iterations. However, the hyperparameters $s$ and $\Delta$ are quite decisive in MARS and we optimize these two scalar values with neuro-evolution \cite{otte2016} for 50 iterations. All the other hyperparameters are adapted manually without any grid search as in the previous works.

\begin{table}[h]
\centering
\caption{Total training time in seconds. Results are taken from \cite{walker2024logncde}. MARS is usually hundreds of times faster than its counterparts.}
\footnotesize
\label{tab:training_time}
\ra{1.2}
\begin{tabularx}{0.6\linewidth}{Xcccccc}
\toprule
 \textbf{Model} & Worms & SCP1 & SCP2 & Ethanol & Heartbeat & Motor \\
\midrule
\textbf{NRDE}     & 8,402  & 1,947 & 2,331 & 2,932 & 13,927 & 14,166 \\
\textbf{NCDE}     & 27,055 & 1,595 & 1,501 & 3,015 & 2,049  & 4,685  \\
\textbf{Log-NCDE} & 3,365  & 2,260 & 992  & 3,413 & 1,587  & 1,153  \\
\textbf{LRU}      & 1,384  & 139  & 132  & 195  & 100   & 846   \\
\textbf{S5}       & 478   & 224  & 227  & 114  & 208   & 271   \\
\textbf{Mamba}    & 2,321  & 281  & 1,256 & 4,134 & 605   & 726   \\
\textbf{S6}       & 2,465  & 76   & 168  & 102  & 78    & 580   \\
\textbf{MARS}     & \textbf{1.6}   & \textbf{1.4}  & \textbf{0.4}  & \textbf{0.3}  & \textbf{0.2}   & \textbf{0.8}   \\
\bottomrule
\end{tabularx}
\end{table}
\begin{table}[t!]
\caption{Number of trainable parameters for all considered models on the long-range datasets. Results are taken from \cite{rusch2025oscillatory}. Some results are omitted for simplicity, as they exhibit the same trend.}
\footnotesize
\label{tab:params}
\ra{1.2}
\begin{tabularx}{\textwidth}{XCCCCCCC}
\toprule
\textbf{Dataset} & Log-NCDE & LRU & S5 & Mamba & S6 & \mbox{LinOSS-IMEX} & MARS \\
\midrule
\textbf{Worms}     & 37,977  & 101,129 & 22,007  & 27,381   & 15,045 & 26,119 & \textbf{1,505} \\
\textbf{SCP1}      & 91,557  & 25,892  & 226,328 & 184,194  & 24,898 & 447,944 & \textbf{801} \\
\textbf{SCP2}      & 36,379  & 26,020  & 5,652   & 356,290  & 26,018 & 448,072 & \textbf{701} \\
\textbf{Ethanol}   & 31,452  & 76,522  & 76,214  & 1,032,772 & 5,780  & 70,088 & \textbf{1,604} \\
\textbf{Heartbeat} & 168,320 & 338,820 & 158,310 & 1,034,242 & 6,674  & 29,444 & \textbf{401} \\
\textbf{Motor}     & 81,391  & 107,544 & 17,496  & 228,226  & 52,802 & 106,024 & \textbf{801} \\
\bottomrule
\end{tabularx}
\end{table}

\section{Conclusion}
This work presents a significant step in advancing the memristive-friendly reservoir networks. Currently models (such as MF-ESN) are already highly efficient as they are not gradient-based. However, until now, the recurrent computation was implemented only sequentially and this limits scaling as the computation complexity increases linearly with increased temporal dimension and each stacked layer. In MARS, we kept the memristive dynamics equivalent to MF-ESN while parallelizing the recurrent dynamics. This improvement in efficiency, led to an advanced scaling behavior. Furthermore, a deeper analysis of memristive-dynamics and the subsequent integration of subtractive skip connections have unleashed the potential of memristive computation in classification tasks. Our results show that this parallel variant not only achieves substantial runtime improvements but also enhances predictive performance even in comparison to large scaled SoTA models. These findings align with the broader trend in sequential modeling that emphasizes the utility of linear recurrent structures when embedded within nonlinear and hierarchical architectures. Crucially, MARS maintains compatibility with the physical intuition behind memristive computation, making it a compelling candidate for future neuromorphic systems. Looking ahead, we anticipate that hybrid approaches, combining bio-inspired computation with scalable algorithmic design, will play a crucial role in shaping the next generation of energy-efficient, high-performance deep learning systems. Furthermore, MARS has plenty of hyperparameters and only a few learnable parameters. The hyperparameters are currently selected manually by try-and-error. A suitable hyperparameter optimization strategy could help improve the performance.

\bibliographystyle{unsrtnat}
\bibliography{references}

\newpage
\appendix

\section{Further information about selected datasets}
\begin{table}[h]
\caption{Overview of the datasets used in the classification experiments in \autoref{sec:mars_in_res}. Reported information includes: the number of sequences in training (Train Size) and test (Test Size), the max length of a sequence in the dataset (Length), the number of output classes (Classes) for the time-series classification problems, the number of input features (Input Dim.) and the type of sequences (Type). }\label{tab:datasets}
\footnotesize
\ra{1.2}
\begin{tabularx}{\linewidth}{lCCCCCC}
\toprule
\textbf{Dataset}\hspace{1.2cm} & Train Size\hspace{0.1cm} & Test Size\hspace{0.1cm} & Length & Classes & Input Dim. & Type\\
\midrule
\textbf{Epilepsy} & $137$ & $138$ & $207$ & $4$ & $3$ & HAR\\
\textbf{SyntheticControl} & $300$ & $300$ & $60$ & $6$ & $1$ & Simulated\\\textbf{GunPoint} & $50$ & $150$ & $150$ & $2$ & $1$ & HAR\\
\textbf{ECG5000}   & $500$ & $4,500$ & $140$ & $5$ & $1$ & ECG\\
\textbf{Coffee} & $28$ & $28$ & $286$ & $2$ & $1$ & Spectro\\
\textbf{JapaneseVowels}  & $270$ & $370$  & $29$ & $9$ & $12$ & Audio\\
\textbf{Wafer}  & $1,000$ & $6,164$ & $152$ & $2$ & $1$ & Sensor\\
\bottomrule
\vspace{0.01cm}
\end{tabularx}

\end{table}

\begin{table}[h]
\caption{We report the best hyperparameter combinations for MARS for each classification task in \autoref{sec:mars_in_res}. The given parameters were found manually by try and error. $\omega$ is the input scaling, $\beta$ is the bias scaling. For the hyperparameters of ESN and MF-ESN, please refer to \cite{pistolesi2025memristive}.}
\label{tab:hypers}
\ra{1.2}
\footnotesize
\begin{tabularx}{\linewidth}{XCCCCC}
\toprule
\textbf{Dataset} & ~~~~~~~~$\omega$ & $\beta$ & $\gamma$ & $\Delta$ & $s$ \\
\midrule
\textbf{Epilepsy}          & ~~~~~~~~0.1 & 0.1 & 1.0 & 0.05 & 5 \\
\textbf{SyntheticControl}  & ~~~~~~~~0.1 & 0.5 & 1.0 & 0.1  & 6 \\
\textbf{GunPoint}          & ~~~~~~~~0.1 & 0.1 & 1.0 & 0.1  & 6 \\
\textbf{ECG5000}           & ~~~~~~~~0.1 & 0.2 & 1.0 & 0.08 & 6 \\
\textbf{Coffee}            & ~~~~~~~~0.1 & 0.1 & 1.0 & 0.1  & 5 \\
\textbf{JapaneseVowels}    & ~~~~~~~~0.1 & 0.1 & 1.0 & 0.5  & 1 \\
\textbf{Wafer}             & ~~~~~~~~0.1 & 0.1 & 1.0 & 0.05 & 14 \\
\bottomrule
\end{tabularx}
\end{table}

 \end{document}